\documentclass[11pt]{article}
\usepackage[margin=1in]{geometry}
\usepackage{amsmath}
\usepackage{amssymb}
\usepackage{graphicx}
\usepackage{xcolor}
\usepackage{url}
\usepackage{array}
\usepackage{caption}
\usepackage{caption}
\usepackage{titlesec}
\usepackage[authoryear,round]{natbib}

\titlespacing*{\section}{0pt}{1.2\baselineskip}{0.45\baselineskip}
\titlespacing*{\subsection}{0pt}{0.85\baselineskip}{0.25\baselineskip}
\titlespacing*{\subsubsection}{0pt}{0.65\baselineskip}{0.2\baselineskip}

\begin{document}

\begin{center}
  \LARGE\textbf{Temporal Coverage over Density: Parsimonious Training-Set Design for ML Climate Downscaling}\\[0.5em]
  \normalsize Karandeep Singh\(^{1}\), Stefan Rahimi\(^{2}\), Chad W. Thackeray\(^{1}\), Stephen Cropper\(^{1}\), Alex Hall\(^{1}\)\\[0.5em]

  \small \(^{1}\)Atmospheric and Oceanic Sciences, University of California, Los Angeles, CA, USA\\
  \small \(^{2}\)Department of Atmospheric Science, University of Wyoming, WY, USA\\[0.5em]

\end{center}

\begin{abstract}
  High-resolution regional climate simulations provide critical information for climate impacts assessments but remain computationally expensive, motivating the development of machine-learning downscalers and emulators. A key challenge is determining how limited high-resolution simulations should be distributed across a changing climate trajectory to capture both forced climate response and internal variability. Using the CESM2 Large Ensemble over the western United States, we compare three training-year selection strategies under fixed data budgets: a contiguous block of historical years, years drawn from both the beginning and end of the simulation period, and years distributed throughout the full climate trajectory. Including both historical and future years consistently outperforms training on historical years alone, demonstrating the importance of exposing downscaling models to climate states outside the historical record and highlighting limitations of stationarity assumptions common in statistical downscaling. Training on years distributed throughout the full climate trajectory performs best overall, indicating that broad sampling of internal variability provides additional information beyond exposure to the forced climate response alone. Models trained on temporally distributed subsets more successfully reproduce variability in unseen ensemble members while retaining strong performance across a wide range of climate diagnostics. Even when trained on only one-tenth of the available high-resolution years, temporally distributed models remain highly competitive with full-data training. These results suggest that, under fixed computational budgets, broad sampling of climate states is more valuable than temporal continuity when allocating scarce high-resolution simulations. The findings provide practical guidance for regional climate downscaling and large-ensemble projection workflows.
\end{abstract}

\section{Introduction}
Climate change introduces persistent non-stationarity into the climate system, directly challenging a core assumption behind many supervised ML workflows: that training and deployment data are sampled from similar distributions~\citep{kouw2019review}. In climate downscaling and surrogate modeling, models are frequently trained on a limited historical interval and evaluated on later periods with shifted thermodynamic baselines, circulation statistics, and extreme-event behavior. Even when average performance is strong, such performance can obscure regime-dependent failure modes, particularly when accuracy is aggregated across periods, regions, or conditions in which the model is more stable.

This issue is particularly important in climate downscaling settings, where models are used not only to produce higher-resolution estimates from low-resolution (LR) climate fields, but also to support scientific inference and decision-relevant diagnostics. In climate applications, users care about physically meaningful behaviors across long horizons: changes in means, shifts in variability, trend fidelity, and extremes. A model that performs well on near-stationary test subsets may still fail under realistic deployment shifts if temporal regime coverage is poor. As a result, dataset design becomes a first-order modeling choice rather than a secondary implementation detail.

At the same time, the data needed to improve temporal robustness are expensive to produce~\citep{hess2025fast}. High-resolution (HR) regional simulations-which provide supervised targets for ML downscaling-are computationally intensive, storage-heavy, and operationally costly over multi-decadal spans. In many workflows, HR generation is the dominant bottleneck. Consequently, practitioners must decide how to allocate a fixed simulation budget across years: densely sample a short interval, or sparsely sample a broader range of climate regimes.

Most prior optimization in climate ML has focused on architecture, loss functions, and training heuristics. By comparison, the temporal composition of the training set has received limited systematic attention~\citep{rampal2024enhancing}, despite its direct connection to distribution shift and compute cost. This paper therefore takes a data-centric view: given limited HR supervision, how should expensive training years be sampled along a non-stationary climate trajectory?

Concretely, we study whether temporal coverage matters more than temporal density in training a performant emulator. We compare three naming-consistent training-year selection strategies: historical contiguous, which uses the earliest available years; historical+future, which combines early and late years; and slice-\(n\), which uses a regular temporal stride. For all three strategies, the total amount of training data is determined by the shared temporal subsampling factor \(N\), where larger \(N\) means fewer selected years; for the slice-\(n\) strategy, the stride \(n\) is then chosen to satisfy that subsampling level. The goal is to determine whether sparse but temporally diverse supervision can preserve downscaling skill while substantially reducing the amount of HR data required.

Our results show three consistent patterns. First, training sets that span broader climatic regimes generalize better than historical contiguous training. Second, temporally sparse sampling can match or nearly match the performance of denser training in many settings. Third, even aggressive sparsification remains competitive, implying large reductions in HR simulation requirements without large losses in predictive skill. Together, these findings suggest a practical principle for ML-driven climate downscaling: prioritize temporal diversity in training-year selection to improve robustness under non-stationarity while controlling end-to-end computational cost.



\section{Problem Formulation}
We consider a supervised climate downscaling setting in which the input is a set of LR climate fields and the target is the corresponding high-resolution (HR) fields over the same domain and time index. Let \(x_t\) denote LR predictors at time \(t\), and let \(y_t\) denote the HR target. A model \(f_\theta\) is trained to learn the mapping \(f_\theta: x_t \mapsto y_t\) by minimizing prediction error on paired LR-HR samples. In practice, this mapping is learned from a finite subset of years, while evaluation and deployment may target later periods, different simulation runs, or external datasets that reflect distinct climate conditions.

The central challenge is distribution mismatch across both time and dataset~\citep{hernanz2024limitations}. If training data are concentrated in a contiguous historical interval from a single source dataset or simulation run, they may insufficiently represent long-term variability, including shifts in means, variances, extremes, and circulation patterns relevant to the prediction task. In practice, this mismatch can become even more pronounced when a model trained on one dataset is deployed to downscale another dataset-even over the same nominal years-because differences in realizations, model physics, bias structure, and internal variability introduce an additional domain shift beyond temporal non-stationarity. Consequently, models can fit the training source well yet generalize poorly when evaluated on different years, different runs, or entirely different datasets. This issue is especially important in realistic downscaling workflows, where the goal is often to train once on an available paired dataset and then apply the learned mapping broadly to other simulations or observationally constrained products.

This problem can be stated as a distribution-shift setting over joint sample-time-dataset tuples:
\[
  P_{\text{train}}(x,y,t,d) \neq P_{\text{test}}(x,y,t,d).
\]
Here, \(d\) indexes the source dataset, simulation run, or domain from which samples are drawn, and \(P_{\text{train}}\) and \(P_{\text{test}}\) denote the joint distributions induced by the selected training years and the deployment dataset/time period, respectively. Train data selection strategies with broader temporal support can produce training data that better cover multiple climate regimes, thereby reducing one important component of train-test mismatch even when evaluation occurs on a different run or dataset. This motivates temporally sparse but distributed sampling as an alternative to dense contiguous sampling. Instead of maximizing temporal density within a short interval, sparse selection across a wider horizon can improve regime coverage per simulated year and lower the amount of expensive HR data required. Under a fixed simulation budget, such designs may offer a better robustness-cost trade-off than conventional historical contiguous training.

\section{Data and Experimental Design}

\subsection{Dataset and Variables}
All climate data for this work are derived from the WUS-D3 dataset~\citep{rahimi2024overview}, a high-resolution regional climate projection dataset for the western United States. WUS-D3 was generated by dynamically downscaling CMIP6 global climate model outputs using the Weather Research and Forecasting (WRF) model.
We use a dynamically downscaled ensemble of Earth system model (ESM) simulations from the Community Earth System Model version 2 Second Large Ensemble (CESM2-LE)~\citep{rodgers2021ubiquity, danabasoglu2020community}. Specifically, 10 ensemble members were one-way dynamically downscaled to 45- and 9-km grids using the Weather Research and Forecasting (WRF) model. Before downscaling, the ESM boundary conditions were bias adjusted~\citep{bruyere2014bias} following the same procedure used for the ESMs downscaled in~\citep{rahimi2024understanding}. The 10 members are structurally similar, with differences arising from distinct sequences of weather and internal variability. Run 1011 is one such initial-condition variant within the large ensemble, whose integrations begin in 1850; because of computational chaos, the sequencing of weather diverges substantially after roughly two weeks of numerical integration. Our training data span 1980-2100. We use the \textit{real} 45-km data as the LR input and the dynamically downscaled 9-km data as the HR target.

The primary predicted variables and input predictors are summarized in Table~\ref{tab:variables}. Each pressure level and vector component is treated as a separate input variable, resulting in 22 input channels. For train set construction, samples are selected from CESM2-LE run 1011 under the temporal sampling strategies described below, while testing is performed on a different dynamically downscaled run (for results in this work, we randomly chose run 1051) to assess robustness under distributional differences across realizations.

\begin{table}[t]
  \centering
  \caption{Predicted variables and input predictors used in this study. Pressure levels and components are treated as independent variables, resulting in a total of 22 input variables.}
  \label{tab:variables}
  \begin{tabular}{l p{0.65\linewidth}}
    \hline
    \multicolumn{2}{c}{\textbf{Predicted Variables}} \\
    \hline
    \texttt{prec} & Precipitation \\
    \texttt{t2} & 2-m air temperature \\
    \hline
    \multicolumn{2}{c}{\textbf{Input Predictors}} \\
    \hline
    \texttt{cape} & Convective available potential energy \\
    \texttt{ivt} & Integrated vapor transport; zonal and meridional components \\
    \texttt{prec} & Precipitation \\
    \texttt{q2} & 2-m specific humidity \\
    \texttt{snow} & Snow water equivalent \\
    \texttt{t2} & 2-m air temperature \\
    \texttt{t\_3d} & 3-D temperature at 300, 500, 700, and 850~hPa \\
    \texttt{u\_3d} & 3-D zonal wind at 300, 500, 700, and 850~hPa \\
    \texttt{uv10} & 10-m zonal and meridional wind components \\
    \texttt{v\_3d} & 3-D meridional wind at 300, 500, 700, and 850~hPa \\
    HGT & Static topographic fields \\
    \hline
  \end{tabular}
\end{table}

Preprocessing follows a consistent pipeline across variables and years. We trim 10 grid cells along the latitudinal boundaries and 15 grid cells along the longitudinal boundaries to reduce edge artifacts from the downscaling setup and boundary effects. The LR predictors are then resized to the HR grid size before being passed to the model. For normalization, each variable is z-score scaled by subtracting a mean and dividing by a standard deviation. Statistics are computed with strict split awareness: LR scaling statistics are computed from all LR data, whereas HR scaling statistics are computed only from the HR subset used for training. The same HR statistics are reused to de-scale model outputs during evaluation. This protocol is important for preventing data leakage and enabling fair benchmarking across different temporal sampling strategies.

\subsection{Model Setup}
We use a fixed U-Net-style~\citep{ronneberger2015u} generator together with a PatchGAN-style discriminator~\citep{goodfellow2014generative,  isola2017image} across all experiments. The generator maps a multivariate low-resolution input tensor \(x\in\mathbb{R}^{H\times W\times C_{\mathrm{in}}}\) to a high-resolution prediction \(\hat{y}\in\mathbb{R}^{H\times W\times C_{\mathrm{out}}}\), where \(C_{\mathrm{in}}\) equals the number of input channels (22 in our case, refer Table~\ref{tab:variables})and \(C_{\mathrm{out}}\) equals the number of target channels. A detailed architectural specification is provided in the Supplementary Information (SI).
This shared architecture is held constant for every training-year selection strategy so that observed performance differences can be attributed to temporal data composition rather than model-capacity changes.

In addition to the reconstruction pathway, we use a GAN-based training setup~\citep{stengel2020adversarial} in which the adversarial component is implemented with a Softplus formulation for stable optimization. To keep comparisons focused on temporal sampling effects and to reduce confounding from objective-function design, we use MAE (L1) as the primary reconstruction loss in all experiments. This consistent objective makes the study data-centric: model and loss are fixed, while temporal training subset design is varied. Implementation details can be found in SI section Model Details.

\subsection{Training Data Selection Strategies}
\label{sec:training_data_selection}
Let \(\mathcal{Y}=\{1980,\dots,2100\}\) denote the available training years from CESM2-LE run 1011, with \(T=|\mathcal{Y}|=121\). We train on a subset \(\mathcal{S}\subseteq\mathcal{Y}\), using all days in each selected year, and report data usage by the number of selected years \(|\mathcal{S}|\) relative to \(T\).

For a temporal subsampling factor \(N\in\{1,2,3,5,7,10\}\), we use a single rounding rule:
\[
  |\mathcal{S}|=\left\lfloor\frac{T}{N}\right\rfloor.
\]
For each temporal subsampling factor \(N\), all three sampling strategies use the fixed \(|\mathcal{S}|\), and we compare:
\begin{itemize}
  \item \textbf{historical contiguous:} select the first \(|\mathcal{S}|\) years of \(\mathcal{Y}\), concentrating supervision in the earliest part of the record. For example, when \(N=10\) and \(|\mathcal{S}|=12\), this strategy uses 1980-1991.
  \item \textbf{historical+future:} split the selected years between the beginning and end of the record. Let \(k=\lfloor |\mathcal{S}|/2\rfloor\), so that \(\mathcal{S}=\{\text{first }k\text{ years of }\mathcal{Y}\}\cup\{\text{last }|\mathcal{S}|-k\text{ years of }\mathcal{Y}\}\). For example, when \(N=10\), this strategy uses 1980-1985 and 2095-2100.
  \item \textbf{slice-\(n\):} select years using a regular temporal stride over \(\mathcal{Y}\), with \(n\) chosen under the same temporal subsampling factor \(N\) to match \(|\mathcal{S}|\). For example, when \(N=10\), the slice-10 strategy selects every 10th year, 1980, 1990, \(\ldots\), 2090, yielding 12 training years distributed across the full period.
\end{itemize}

As an upper-bound baseline, we also train on all years from run 1011 (\(N=1\), 100\% usage). For \(T=121\): \(N=1\Rightarrow 121\) years, \(N=2\Rightarrow 60\) years, \(N=5\Rightarrow 24\) years, and \(N=10\Rightarrow 12\) years.

These strategies are designed to separate the effects of exposure to the forced climate change signal from the effects of sampling internal variability across the climate trajectory. Table~\ref{tab:climate_sampling_strategies} summarizes the scientific role of each training-data composition.

\begin{table}[!ht]
  \centering
  \small
  \setlength{\tabcolsep}{4pt}
  \caption{Climate sampling strategies and their interpretation in terms of forced climate change and internal variability.}
  \label{tab:climate_sampling_strategies}
  \begin{tabular}{>{\raggedright\arraybackslash}m{0.22\linewidth} >{\raggedright\arraybackslash}m{0.19\linewidth} >{\raggedright\arraybackslash}m{0.18\linewidth} >{\raggedright\arraybackslash}m{0.29\linewidth}}
    \hline
    \textbf{Sampling strategy} & \textbf{Forced climate change signal} & \textbf{Internal variability} & \textbf{Scientific interpretation} \\
    \hline
    \textbf{Historical only} & Limited & Limited & Baseline representing traditional training on historical climate \\
    \textbf{Historical + future} & Stronger & Limited & Tests the importance of exposing the model to the forced climate trajectory \\
    \textbf{Slice-\(n\)} & Stronger & Stronger & Tests the additional value of sampling a broader range of climate states and variability \\
    \hline
  \end{tabular}
\end{table}

The resulting comparisons isolate complementary aspects of training-set design:
\begin{itemize}
  \item \textbf{Historical+future vs. historical-only:} isolates the value of sampling the forced climate change signal. The superior performance of historical+future suggests that stationarity assumptions matter and that models trained only on historical climate do not fully learn the mapping required in future climates.
  \item \textbf{Slice-\(n\) vs. historical+future:} isolates the value of sampling internal variability across the climate trajectory. The superior performance of slice-\(n\) suggests that broad sampling of climate states provides information beyond exposure to the forced response alone.
  \item \textbf{Slice-\(n\) vs. full training:} quantifies how much information can be retained with sparse sampling of climate states and therefore how efficiently high-resolution simulations can be allocated.
\end{itemize}

\subsection{Evaluation Protocol}
Evaluation is performed on all years (1980-2100) from CESM2-LE runs other than the training run (1011), using fully held-out run realizations. In the main text, we report results for one randomly selected held-out member, run 1051; the pointwise-statistics analysis for a second randomly selected held-out member, run 1031, is provided in the SI. This protocol tests cross-run temporal generalization: although training subsets may cover historical and/or future periods in run 1011, evaluation is conducted on distinct runs across the full time horizon, so model behavior is assessed under unseen realization-level variability and regime evolution.

We report pointwise reconstruction metrics (root mean squared error, RMSE; mean absolute error, MAE; and anomaly correlation coefficient, ACC), climate-impact diagnostics, seasonal linear trends, and spectral diagnostics that assess recovery of high-frequency spatial structure. The model is trained using L1 (MAE) loss, and none of the secondary diagnostics are directly optimized during training; they are used only for out-of-objective evaluation of physical and statistical fidelity. No single diagnostic is sufficient to characterize overall model skill, so we interpret these metrics jointly. This is important because pointwise metrics tend to improve monotonically with larger training-set sizes, whereas trend, extreme, and spectral skill can exhibit more complex behavior~\citep{bano2020configuration}. We therefore evaluate both \texttt{prec} and \texttt{t2} across complementary diagnostics, since their error characteristics and physically relevant structures differ.

\begin{table}[t]
  \centering
  \caption{Evaluation diagnostics for precipitation (\texttt{prec}) and 2-m temperature (\texttt{t2}).}
  \label{tab:eval_metrics}
  \begin{tabular}{l p{0.65\linewidth}}
    \hline
    \multicolumn{2}{c}{\textbf{Precipitation (\texttt{prec})}} \\
    \hline
    RR & Total precipitation (mm) \\
    CWD & Maximum consecutive wet days (days) \\
    CDD & Maximum consecutive dry days (days) \\
    R10mm & Heavy precipitation days ($\ge 10$ mm) (days) \\
    RX1day & Maximum 1-day precipitation (mm) \\
    RX5day & Maximum 5-day precipitation (mm) \\
    \hline
    \multicolumn{2}{c}{\textbf{2-m Temperature (\texttt{t2})}} \\
    \hline
    WSDI & Warm-spell duration index (days) \\
    TX10p & Days with TX below the 10th percentile (days) \\
    TX99p & Days with TX above the 99th percentile (days) \\
    CSDI & Cold-spell duration index (days) \\
    TXx & Maximum daily mean temperature ($^\circ$C) \\
    TXn & Minimum daily mean temperature ($^\circ$C) \\
    \hline
  \end{tabular}
\end{table}

\section{Results}

\subsection{Effect of Temporal Coverage on Generalization}
This section evaluates how the temporal placement of HR training years affects generalization under a fixed model and evaluation setup. We first examine whether temporal placement matters when the model, objective, and evaluation protocol are fixed. Overall aggregate skill remains good across training-year selection strategies~\citep{bano2021suitability}, but historical contiguous training consistently underperforms strategies that span broader climate regimes. When training is restricted to early historical years, the model still attains reasonable skill on held-out CESM2-LE runs, yet its robustness is weaker than that of temporally distributed alternatives, consistent with temporal regime mismatch. The historical+future strategy improves robustness by exposing the model to both early and late climate states across the 1980-2100 horizon. The slice-\(n\) strategy is similarly effective, and is particularly attractive because it distributes supervision across the full time axis while also covering intermediate regimes that endpoint-style historical+future selection may miss.

\begin{figure}[t!]
  \centering
  \includegraphics[width=0.94\linewidth]{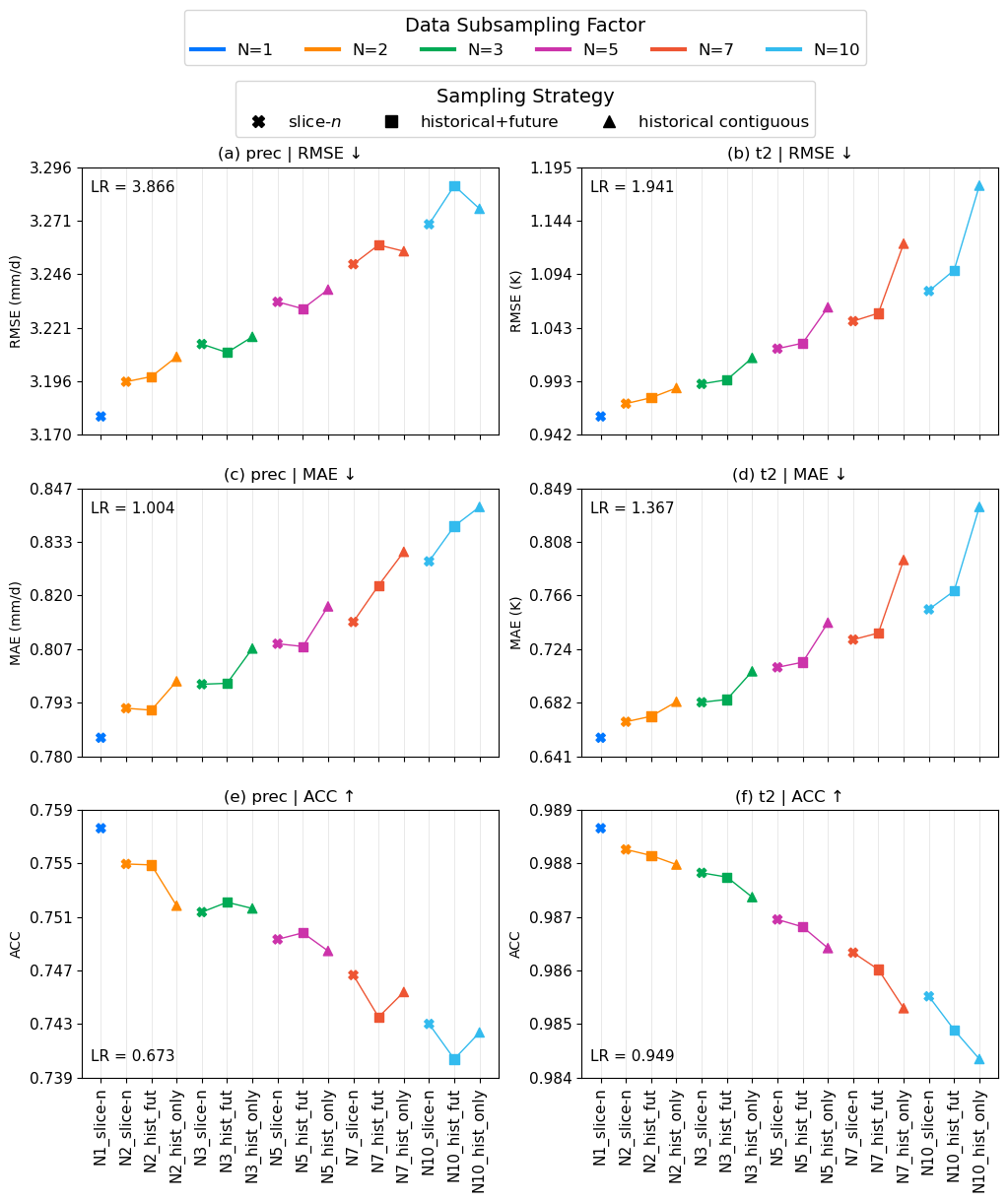}
  \caption{\small{RMSE, MAE, and ACC evaluated at a fixed training epoch (500) across all temporal subsampling settings \(N\) and training-year selection strategies. The LR baseline values are omitted from the plot and reported in the text because their RMSE values fall outside the plotted range of the trained models; including them would require a much wider axis range, reducing visual resolution within the trained-model regime and making cross-configuration differences difficult to interpret. For compactness, x-axis labels use shortened strategy names: historical+future is denoted as \texttt{hist\_fut}, historical contiguous as \texttt{hist\_only}, and each label appends the corresponding temporal subsampling factor \(N\).}}
  \label{fig:pointwise_metrics}
\end{figure}

Figure~\ref{fig:pointwise_metrics} summarizes RMSE, MAE, and ACC at a fixed training epoch. As expected, performance improves with denser training sets (smaller \(N\)), but even under aggressive sparsification, the loss remains modest.
For example, in the slice-10 setting, which uses only 1/10 of the available HR training data, RMSE increases by just 3.13\% for \texttt{prec} and 12.81\% for \texttt{t2} relative to the full-data setting. Thus, even with a tenfold reduction in HR supervision, the model retains competitive skill rather than exhibiting catastrophic or even 10x degradation. Figure~\ref{fig:1031_vs_1051_pointwise_stats} presents these statistics for another randomly selected CESM2-LE realization, 1031, alongside the results from realization 1051 for comparison. Model performance is expected to generalize across different realizations, as supported by the consistent behavior observed in Figure~\ref{fig:1031_vs_1051_pointwise_stats}.

\subsection{Climate Diagnostics Beyond Pointwise Error}

\begin{figure}[t!]
  \centering
  \includegraphics[width=0.95\linewidth]{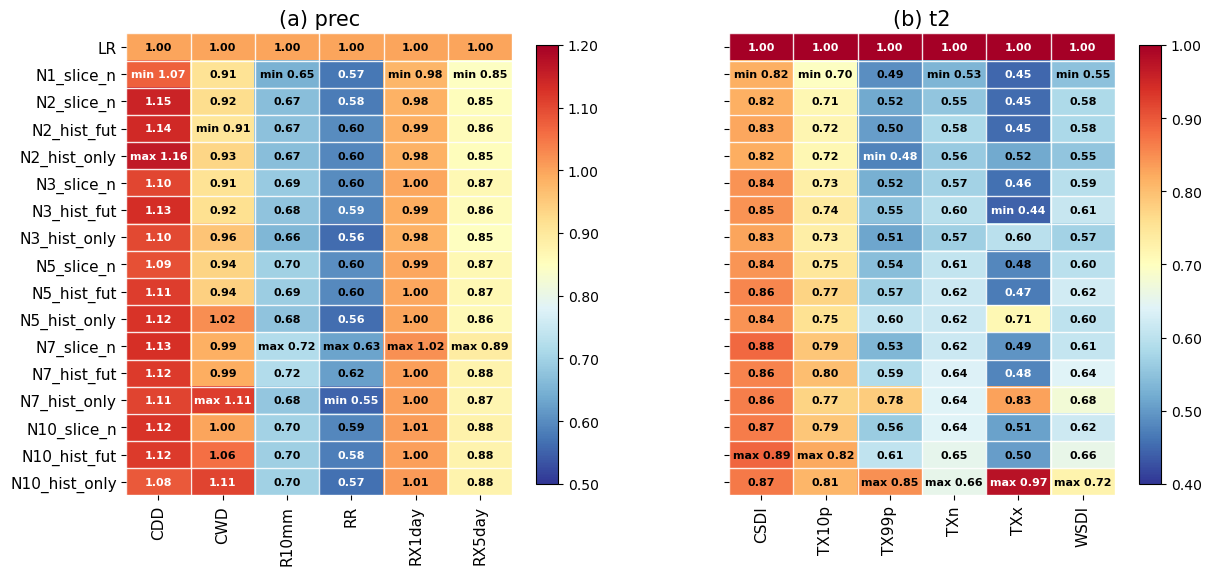}
  \caption{\small{Normalized RMSE for event-based precipitation and temperature diagnostics across temporal subsampling levels (N) and training-year selection strategies. The diagnostics are defined in Table~\ref{tab:eval_metrics}. Values are normalized by the corresponding LR RMSE for each diagnostic, so that the LR baseline is equal to 1. Values below 1 indicate improvement relative to LR, while values above 1 indicate larger error than LR. The minimum and maximum values within each diagnostic are marked for reference. These diagnostics are not directly optimized during model training and therefore provide an out-of-sample assessment of whether the downscaled fields preserve physically relevant precipitation and temperature behavior. Metrics based on event persistence, such as CDD and CSDI, show comparatively weaker performance for several ML configurations, indicating that spell-duration behavior remains a more stringent diagnostic than pointwise RMSE. The y-axis labels are shortened as in Fig.~\ref{fig:pointwise_metrics}}}
  \label{fig:prec_t2_skill_metrics}
\end{figure}

\begin{figure}[h!]
  \centering
  \includegraphics[width=0.95\linewidth]{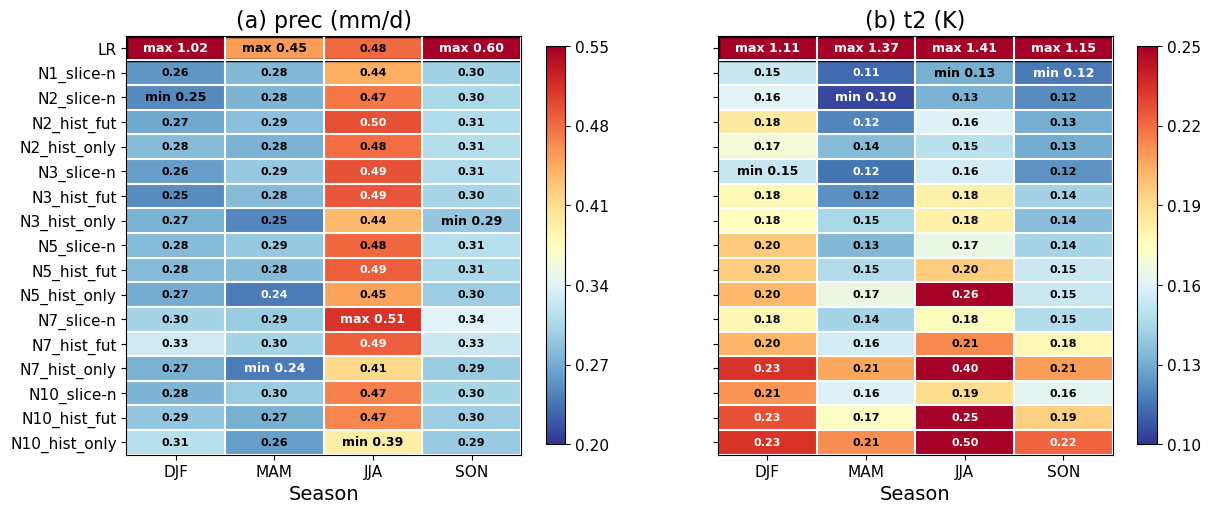}
  \caption{\small{Seasonal climatology RMSE for \texttt{prec} and \texttt{t2} across temporal subsampling levels (N) and training-year selection strategies. For each season, climatologies are computed over the full 1980-2099 period, and RMSE is evaluated relative to the HR reference climatology. The LR row gives the corresponding error of the low-resolution baseline. The minimum and maximum RMSE values within each season are marked for reference. Color scales are shown separately for \texttt{prec} and \texttt{t2} because the two variables have different units and error ranges; colors should therefore be compared within each panel rather than across variables. The y-axis labels are shortened as in Fig.~\ref{fig:pointwise_metrics}.}}
  \label{fig:pointwise_climatology_metrics}
\end{figure}

The pointwise results establish the main robustness pattern, but pointwise error provides only one view of downscaling quality~\citep{maraun2015value}. In many climate applications, the primary scientific requirement is not only accurate grid-cell values, but also the preservation of physically meaningful behavior, including climatological means, seasonal structure, long-term trends, and event-based diagnostics. Figure~\ref{fig:prec_t2_skill_metrics} therefore evaluates the broader set of climate diagnostics introduced in Table~\ref{tab:eval_metrics}~\citep{zhang2011indices}. Consistent with the pointwise analysis, most ML configurations improve on LR for many precipitation and temperature diagnostics, including total precipitation, heavy-precipitation frequency, precipitation extremes, warm-spell duration, percentile-based temperature exceedances, and daily temperature extrema. The benefit is particularly notable for temporally distributed sampling strategies, which retain much of the diagnostic skill even at substantially reduced data budgets..

Nevertheless, the improvement is not universal. CDD and CSDI stand out as diagnostics for which several ML simulations are systematically worse than LR. This indicates that the models can reproduce many aspects of the marginal distribution while still degrading aspects of temporal persistence. Because CDD and CSDI are spell-length metrics, they are sensitive to the ordering and continuity of threshold events rather than only the magnitude of individual daily errors. Small day-to-day biases, occasional threshold crossings, or overly intermittent predictions can therefore shorten or lengthen spells and produce worse RMSE than LR. These results should also be interpreted in light of the experimental objective: the configurations are designed to evaluate how much diagnostic skill can be retained under reduced training-data budgets, rather than to fully optimize model fidelity for every individual climate index.
These results suggest that reduced-data downscaling is broadly effective for many climate diagnostics, but that persistence-based dry and cold-spell behavior remains a more stringent test of physical realism.

\begin{figure}[t!]
  \centering
  \includegraphics[width=0.95\linewidth]{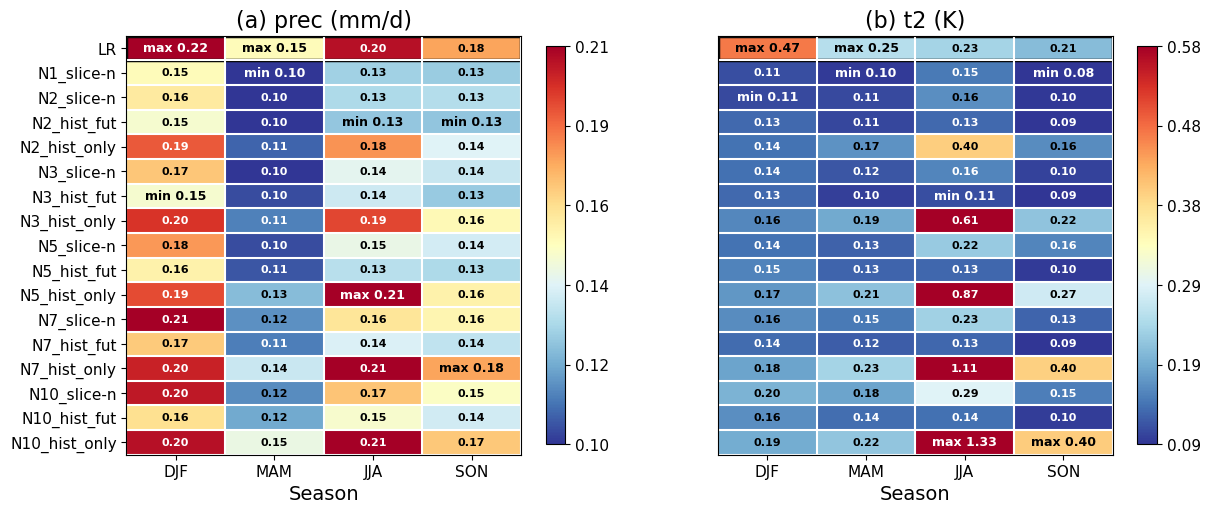}
  \caption{\small{Seasonal trend-slope RMSE for \texttt{prec} and \texttt{t2} across temporal subsampling levels (N) and training-year selection strategies. For each season, a linear trend is fitted over the full 1980-2099 period, and RMSE is computed between the resulting trend slopes from each simulation and the corresponding HR reference trend slopes. The LR row gives the corresponding trend-slope error of the low-resolution baseline. The minimum and maximum RMSE values within each season are marked for reference. Color scales are shown separately for \texttt{prec} and \texttt{t2} because the fitted trend slopes have variable-specific units and error ranges; colors should therefore be compared within each panel rather than across variables. The y-axis labels are shortened as in Fig.~\ref{fig:pointwise_metrics}}}
  \label{fig:seasonal_trend_rmse}
\end{figure}



Figure~\ref{fig:pointwise_climatology_metrics} presents the seasonal climatologies together with their associated RMSE values, while Figure~\ref{fig:seasonal_trend_rmse} reports the corresponding errors in seasonal linear trends for \texttt{prec} and \texttt{t2}. Across both diagnostics, temporally sparse training retains the dominant climatological structure and preserves meaningful trend fidelity, with only modest degradation relative to denser training configurations. Climatological averaging suppresses high-frequency variability and, under this view, performance differences between denser and sparser training settings narrow further.

The largest climatology errors occur in JJA, particularly for \texttt{t2} and, to a lesser extent, \texttt{prec}. This seasonal dependence is consistent with the magnitude of the required HR-LR climatological correction in the training data. We quantify this correction as the seasonal climatological difference between the high-resolution target and the low-resolution input, after both fields are evaluated on the analysis grid. For precipitation, the required HR-LR correction is small in DJF and MAM, about 0.04 mm/d in both seasons, but increases to 0.27 mm/d in JJA and 0.13 mm/d in SON. For \texttt{t2}, the corresponding correction magnitudes are 0.18 K in DJF, 0.36 K in MAM, 0.42 K in JJA, and 0.22 K in SON. Thus, JJA requires the largest climatological adjustment from the LR input toward the HR target for both variables, making it the most demanding season for the downscaling model. The relatively elevated JJA RMSEs should therefore be interpreted in the context of this larger underlying resolution-dependent discrepancy, rather than as an isolated failure mode. In addition, because the same model architecture is used for both \texttt{prec} and \texttt{t2}, the coupled structure of the two targets may also contribute to the seasonal pattern of errors: seasons in which both variables require larger HR-LR corrections may place stronger demands on the model's shared representation. Importantly, even in JJA, the ML configurations generally remain closer to HR than the LR baseline, indicating that the models recover a substantial fraction of the missing seasonal structure.

The trend-slope RMSE shows a broadly similar seasonal structure, with relatively larger errors in seasons where the HR-LR climatological correction is larger. This diagnostic is useful as a compact summary of whether long-term seasonal changes are preserved, but it should be interpreted with some caution. Errors in fitted slopes depend not only on the mean downscaling error, but also on the magnitude of the underlying trend, internal variability, and the length of the fitting period. We therefore treat the seasonal trend RMSE as an aggregate consistency check on long-term behavior, rather than as a complete characterization of forced-trend fidelity.

\begin{figure}[t!]
  \centering
  \includegraphics[width=0.965\linewidth]{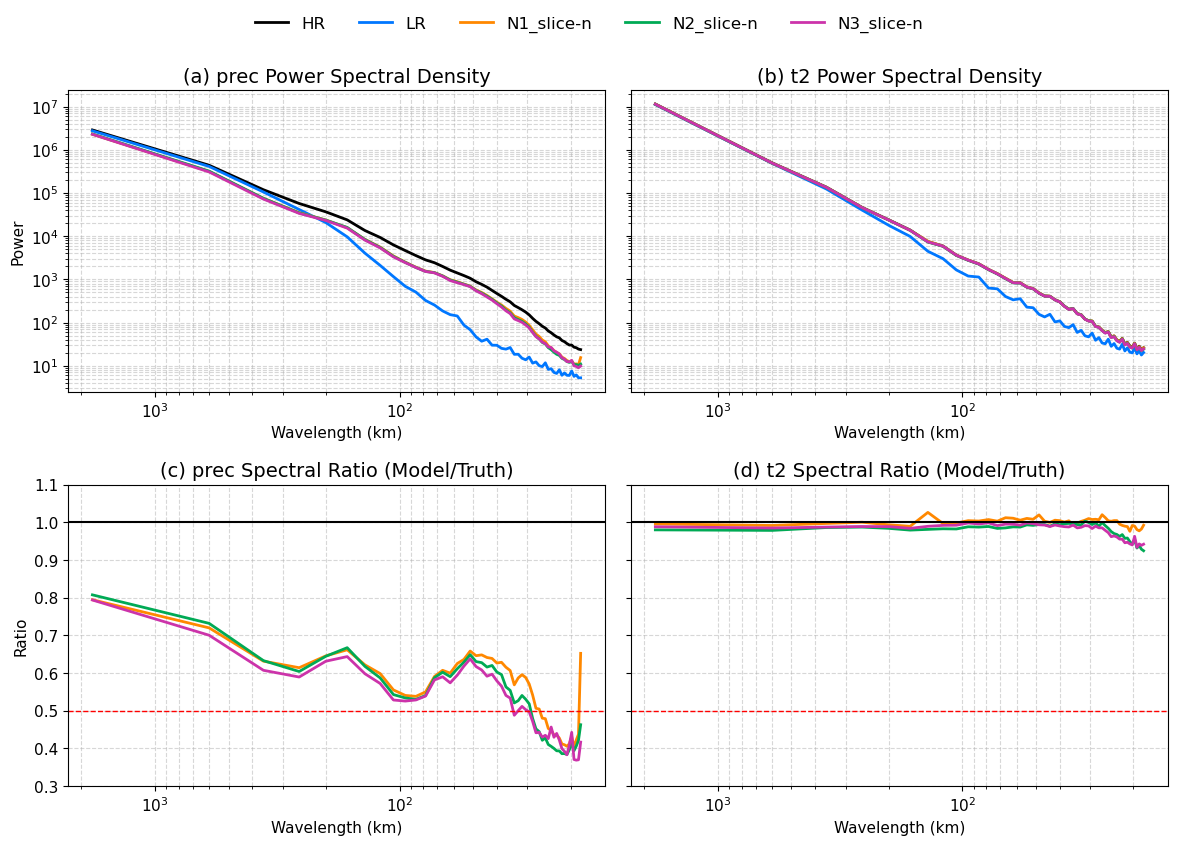}
  \caption{\small{Spectral diagnostics for \texttt{prec} and \texttt{t2} under the slice-\(n\) training-year selection strategy. To improve visualization, only the \(N\in\{1,2,3\}\) temporal subsampling settings are shown. For \texttt{prec}, denser HR training subsets better recover high-frequency variability, although differences are small. The slight upturn at the shortest wavelengths occurs in the low-power spectral tail and is likely sensitive to noise-floor effects rather than a physically meaningful recovery of additional variance. All models underrepresent the lowest-frequency components. For \texttt{t2}, spectral fidelity remains strong across subsampling settings. The legend labels are shortened as in Fig.~\ref{fig:pointwise_metrics}.}}
  \label{fig:spectral_analysis_1_2_3}
\end{figure}

The spectral diagnostics provide a complementary view of spatial-scale fidelity. Whereas the climatology and trend diagnostics summarize seasonally aggregated behavior, Figure~\ref{fig:spectral_analysis_1_2_3} evaluates whether the downscaled fields reproduce the spatial distribution of variance across scales. Across all training-year selection strategies, the ML downscaled outputs largely preserve both low- and high-frequency structure and consistently outperform the LR baseline, especially for mid and higher frequencies. Models trained with denser subsets retain more high-frequency power than those trained on sparser subsets, indicating improved fidelity to fine-scale variability as training coverage increases.

Taken together, these diagnostics reinforce the pointwise results: in nearly all analyses, the ML downscaled outputs substantially outperform the LR inputs. Within the model comparisons, the benefit of temporal coverage remains evident, with historical contiguous training consistently producing the weakest results, despite preserving year-to-year continuity in the climate trends. Although the HR training targets are bias-corrected, no post-hoc bias correction is applied to the ML predictions, so these gains reflect the learned model behavior itself rather than any additional adjustment at inference time.

\subsection{Robustness-Efficiency Trade-off Under Reduced HR Budgets}
The diagnostic results show that temporal coverage can preserve scientific skill; the next question is how much HR data this can save. Our results show that the downscaling model maintains strong skill even when trained with temporally sparse HR supervision, with direct implications for practical dataset design. In climate downscaling pipelines, HR simulations are often the dominant bottleneck: they are computationally expensive, slow to generate, and operationally demanding to store, preprocess, and quality-control. Reducing the number of simulated HR years therefore translates immediately into lower end-to-end project cost.

This effect is substantial even at moderate sparsification. For example, slice-2 uses roughly 50\% of available HR years and thus implies an approximate 50\% reduction in HR simulation burden. More aggressive settings provide larger savings: slice-5 (about 20\% of HR years) can reduce required HR generation by approximately 80\%, while still retaining competitive predictive skill in our experiments. Even with slice-10 (about 10\% of HR training years), the ML downscaling model remains clearly superior to the LR baseline. The key practical point is that the performance drop is much smaller than the data-cost reduction, producing a favorable efficiency regime for many applications.

Temporal sparsification can also improve the ML development stage itself. Because the model maintains performance with fewer training samples, training and hyperparameter sweeps become faster, reducing GPU-hours and shortening turnaround time for iterative experimentation. As a result, the overall workflow can benefit twice: first through reduced dynamical downscaling cost, and second through lower ML training overhead. These gains compound across the full pipeline, from generating regional HR targets to fitting and validating the ML downscaler.

\begin{figure}[t!]
  \centering
  \includegraphics[width=0.95\linewidth]{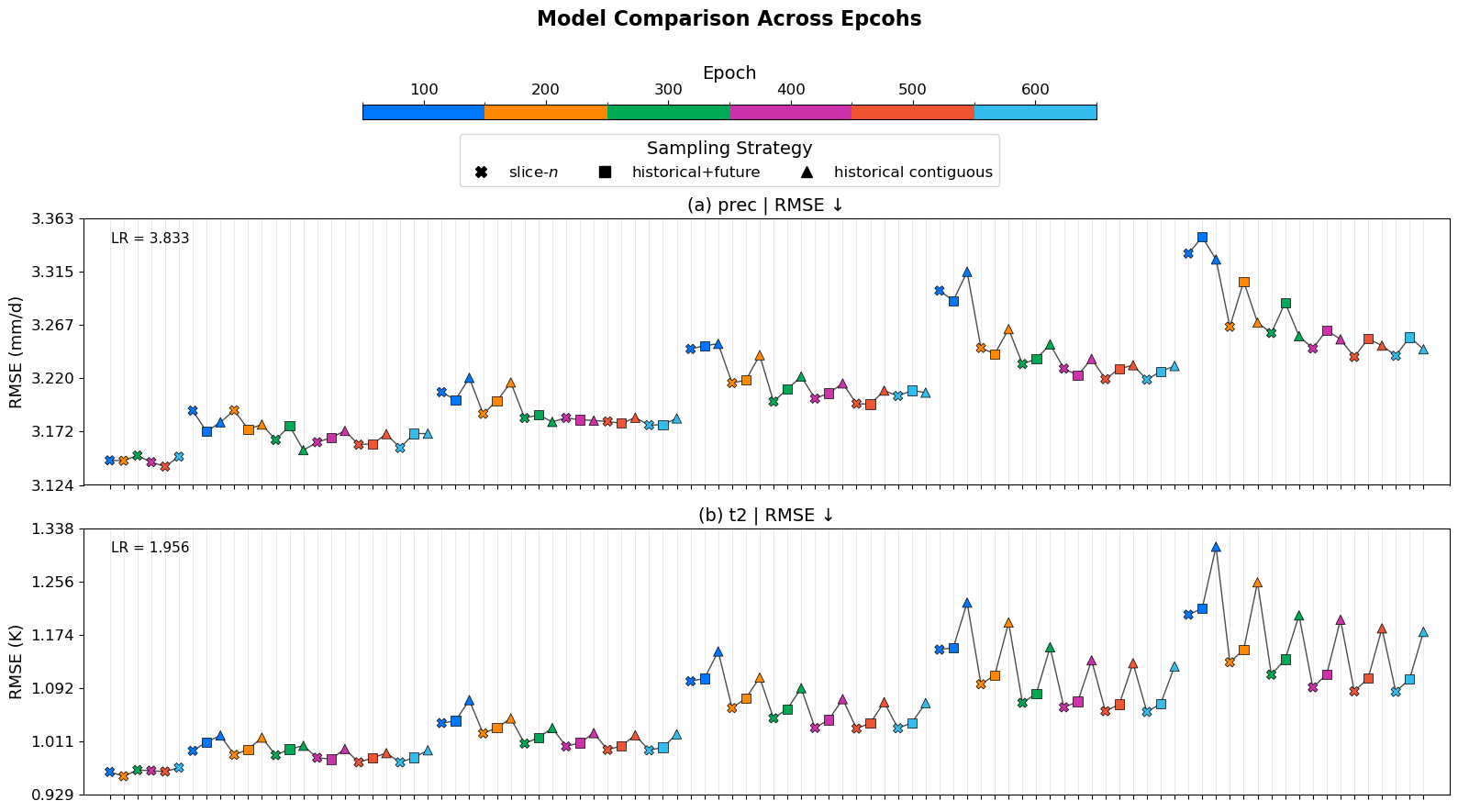}
  \caption{\small{Effect of training duration on performance across training-year selection strategies. Model skill generally improves with additional training, and the slice-\(n\) strategy consistently outperforms the alternative sampling designs. The figure also highlights the interaction between training-year selection strategy and temporal subsampling setting \(N\). Results are shown for training epochs 100-600 in increments of 100. Because the number of optimization steps per epoch varies with the amount of training data, comparisons at fixed epoch counts should be interpreted with this difference in effective training iterations in mind.}}
  \label{fig:models_epochs_types}
\end{figure}

Training duration interacts with this comparison and is important for interpreting fixed-epoch summaries. Figure~\ref{fig:models_epochs_types} shows model performance at epochs 100 through 600 across temporal subsampling settings \(N\) and training-year selection strategies. More training generally leads to better performance, and slice-\(n\) generally outperforms the alternative strategies, especially once training has stabilized. Because the number of optimization steps per epoch varies with the amount of training data, fixed-epoch comparisons should be interpreted with this difference in effective training iterations in mind. For instance, training the model for 100 epochs with slice-1 setting (full data) in our setup results in about 17k iterations; to have the same number of training iterations, the slice-2 setting would need about 200 epochs, and slice-10 would need about 1000 epochs.

Under fixed or limited computational budgets, these findings suggest a pragmatic allocation strategy: prioritize temporal coverage first, then add density as resources permit. This yields an efficient middle ground between cost and expected outcomes, where moderate sparsification preserves most skill gains while substantially lowering HR simulation burden and downstream ML training cost.

\subsection{Why Distributed Sampling Works}
The performance and efficiency results point to the same underlying mechanism: temporal regime coverage~\citep{doury2023regional}. Sampling strategies that distribute training years across the full horizon expose the model to a wider range of climate states, including both background variability and long-term forced changes. This reduces temporal redundancy relative to historical contiguous sampling, where many adjacent years contain similar large-scale conditions and therefore contribute limited new information per simulated year.


\begin{figure}[h!]
  \centering
  \includegraphics[width=0.99\linewidth]{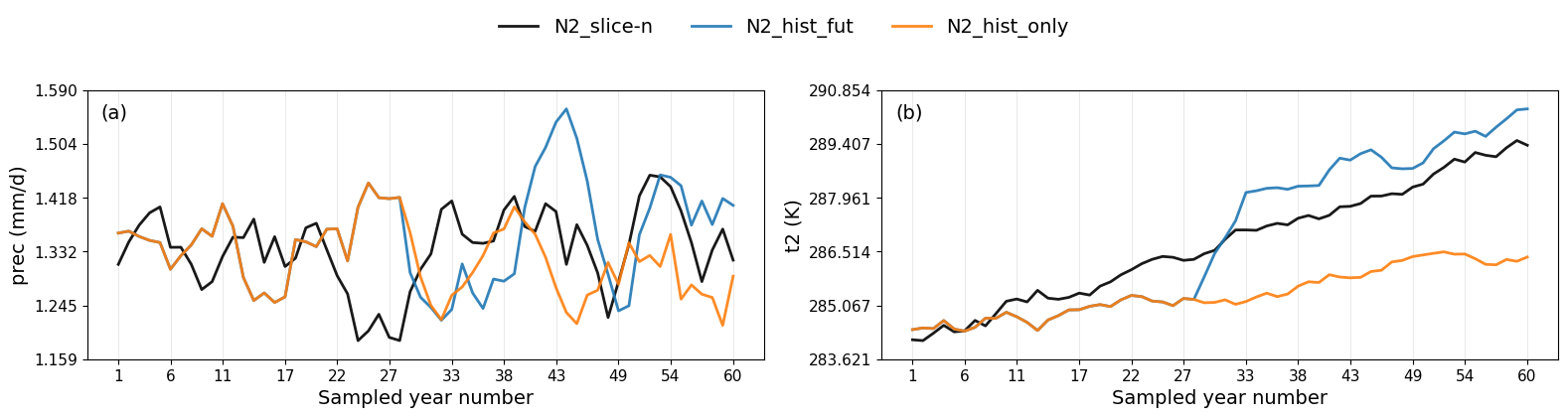}
  \caption{\small{Annual means of \textit{HR training data} for \texttt{prec} and \texttt{t2} at temporal subsampling factor \(N=2\), using about 50\% of available years. For \texttt{t2}, a clear warming trend is visible in the slice-2 and historical+future training sets, whereas the historical contiguous training set does not capture this warming signal. For \texttt{prec}, trends are relatively flat. The historical contiguous and historical+future curves overlap over the shared historical interval because both strategies use the same years in that range. To improve visualization, the values shown here are moving averages computed using a window size of 5. The legend labels are shortened as in Fig.~\ref{fig:pointwise_metrics}}}
  \label{fig:hr_prec_t2_trends}
\end{figure}

A second mechanism is improved alignment between training and evaluation distributions under non-stationarity. When training includes broader temporal support, the model sees a more representative span of target behavior and can better generalize to held-out periods and runs. This interpretation is also consistent with time-of-emergence work showing that forced climate signals become increasingly distinguishable from internal variability over time~\citep{hawkins2012time}. In our case, this is consistent with the broader target-statistic coverage induced by distributed sampling. Yearly means and related diagnostics under slice-\(n\) designs span a wider portion of the century-scale distribution than historical contiguous subsets. This pattern is visible in the \textit{HR training data} annual means in Figure~\ref{fig:hr_prec_t2_trends}, where \texttt{t2} shows a clear warming signal captured by slice-2 and historical+future sampling but missed by historical contiguous sampling. In contrast, the \texttt{prec} annual-mean trends are relatively flat, indicating that all sampling strategies perform similarly for precipitation trend representation.

ML model capacity and regularization also matter for realizing these benefits. If model capacity is too low, it may fail to learn key climate relationships and will underfit even with diverse data. If capacity is too high without sufficient regularization, sparse training can lead to rapid overfitting and poor out-of-regime performance. The model used here has moderate capacity (about 38M parameters). This balance appears important for maintaining skill under sparse supervision.

The competitive performance of the historical+future strategy relative to slice-\(n\) is also mechanistically consistent. Over 1980-2100, the CESM2-LE trajectory exhibits a persistent warming signal and associated distributional shifts that are especially pronounced toward late-century conditions (see Figure~\ref{fig:hr_prec_t2_trends}). Historical+future sampling places training points in both early and shifted late regimes, while slice-\(n\) sampling spreads coverage throughout the transition. Both strategies therefore reduce train-test mismatch compared with historical contiguous training, which remains concentrated in one side of the evolving distribution.

Overall, these mechanisms point to the same conclusion: temporally diverse supervision improves robustness under climate non-stationarity by increasing regime coverage, reducing redundant sampling, and enabling better use of model capacity under fixed compute constraints.

\section{Discussion}
\subsection{Practical Implications}
The results above translate directly into design guidance for expensive regional simulation campaigns. In ML-driven downscaling workflows, HR target generation is typically the dominant cost in compute, storage, and human effort. Under this constraint, temporally sparse supervision provides a strong efficiency lever: even slice-2 corresponds to roughly a 50\% reduction in HR years. Importantly, even slice-10 does not produce a collapse in model behavior; instead, the model retains useful fidelity in trend- and diagnostic-oriented evaluations. This indicates that a simple change in year-selection strategy can substantially reduce end-to-end budget for both HR simulation and ML training.

For reference, on our hardware (with 4 A100 80G GPUs), data from slice-2 onward can be loaded into memory. Training slice-2 for 600 epochs takes approximately 8 hours 17 minutes, whereas slice-3 requires about 5 hours 22 minutes. For slice-1 training, and for inference over all data points in a CESM run, the data must be read from disk; under this configuration, training for 600 epochs takes about 4 days. Despite this disk I/O requirement, inference remains fast, completing in about 8.5 minutes on average. This corresponds to a downscaling throughput of about 85 data points per second, or 0.0116 seconds per data point. Without parallelization, a single dynamically downscaled run of CESM takes about 150 days to complete. Therefore inference has a speedup of about 25,000x.

From a workflow perspective, systematic slice-\(n\) sampling is straightforward to implement and operationally feasible. It does not require specialized infrastructure beyond standard data-subsetting logic, yet it can materially improve robustness-cost trade-offs. This is especially relevant for fixed-budget projects where the scientific objective is to preserve climatological behavior, trends, and broad diagnostic fidelity rather than to minimize only pointwise reconstruction error.

\subsection{Limitations and Scope}
These practical implications should be interpreted with several caveats. Optimization length interacts with temporal subsampling factor \(N\), so strict comparisons should consider matched or appropriately scaled optimization budgets in addition to fixed-epoch summaries. Larger historical contiguous subsets (e.g., historical contiguous at \(N=2\), using the first 50\% of years) can also carry enough signal to learn climate behavior, which can narrow differences among strategies depending on the strength of forced trends.

Sparse strategies can miss critical intermediate regimes or rare events if those events are not sampled. The historical contiguous and historic+future strategies may have the advantage of preserving year-to-year continuity in evolving climate trends and associated temporal dependencies. However, slice-\(n\) sampling covers more temporal ground under the same temporal subsampling factor \(N\), and we generally view it as safer than a strict historical+future endpoint design: evenly distributed sampling lowers the risk of leaving unsampled gaps between temporal tails, especially for datasets with long horizons or non-monotonic regime changes. Even so, applications targeting extremes may benefit from hybrid schemes that augment systematic sampling with event-aware year selection.

Scope is another important consideration. We evaluate a downscaling task over the Western United States (WUS), a region with strong geographic heterogeneity (complex topography, coastal influence, rain-shadow effects, and varied hydroclimate regimes). We also test two output variables with different statistical character: \texttt{prec} (highly skewed, intermittent, and nonlinear)~\citep{maraun2010precipitation} and \texttt{t2} (closer to Gaussian-like behavior and generally easier to downscale)~\citep{sha2020deepI, sha2020deepII}. Because conclusions are consistent across these contrasting variables and a broad predictor set (see Table~\ref{tab:si_input_predictors}), we expect the main trends to transfer to related variables, though absolute performance may vary by target and regime.

Methodologically, results are demonstrated with one strong architecture family (U-Net) and on CESM2-LE ensemble runs, where cross-run distributions are related by construction. Therefore, the exact magnitudes of gains may shift with architecture choice, domain, or data source.
Nonetheless, because the intervention changes the temporal support of the training data rather than the model class itself, we expect the qualitative pattern—that temporal diversity improves robustness per unit HR cost—to hold across reasonable architecture substitutions. This expectation is grounded in the mechanism of the method: exposing the model to a broader range of climate states should reduce sensitivity to any single historical or ensemble-specific distribution, whether the predictive backbone is a U-Net or a newer architectures such as transformers or diffusion-based models. Alternative architectures may alter absolute performance levels, sample efficiency, or the size of the observed gains, particularly if they differ in inductive bias or ability to represent long-range spatial dependencies. However, they are less likely to overturn the central finding that temporally diverse training data provide a cost-effective route to improved out-of-period generalization.

\subsection{Future Directions}
Future work should test these principles across additional ESMs, regions, variables, and architecture families, including diffusion-based, transformer-based, and alternative convolutional downscaling models. Such tests should include transfer across structurally different climate model families and evaluation across additional ensemble runs. Promising extensions include adaptive or learned year-selection policies, objective-aware sampling for extremes and rare events, and joint allocation strategies that optimize both HR simulation and ML training budgets. Together, these directions would extend the simple slice-\(n\) heuristic into a broader data-design framework for robust, compute-efficient climate ML.

\section{Conclusion}
This work investigates a data-design question that is central to climate workflows: how should limited high-resolution supervision be allocated under non-stationary climate dynamics? Our results show that temporal training composition is a primary determinant of downstream robustness. In particular, strategies that increase temporal regime coverage consistently outperform historical contiguous sampling, while providing better coverage of intermediate regimes.

A key practical outcome is that strong downscaling skill can be retained even with substantially less HR data, provided that the temporal subsampling factor \(N\) is paired with an appropriate temporal strategy. This creates a favorable robustness-cost trade-off for climate applications where HR simulation generation dominates overall budget. Because the proposed slice-\(n\) sampling is simple, transparent, and easy to implement in existing pipelines, it offers an immediately actionable path to reducing compute, time, and operational burden without major methodological overhead.

The main takeaways are:
\begin{itemize}
  \item \textbf{Temporal diversity is the main design priority:} temporal regime coverage in training matters more than historical contiguous sampling for robust generalization under non-stationarity.
  \item \textbf{Sparse supervision can remain effective:} temporally distributed subsets preserve much of the skill even with reduced HR data requirements, lowering both HR simulation cost and ML training cost.
  \item \textbf{Simple distributed sampling is actionable:} slice-\(n\) sampling is straightforward to deploy, improves over historical contiguous training, and is generally safer than strict historical+future endpoint sampling when intermediate regimes matter.
\end{itemize}

Overall, these findings support a practical principle for ML-driven climate downscaling under fixed compute budgets: prioritize temporally diverse supervision first, then increase density as resources permit. This provides a robust and implementation-friendly route to reliable downscaling under climate non-stationarity.

\section*{Acknowledgments}
The authors thank the WUS-D3 modeling team for developing the datasets and simulations that made this analysis possible. The authors acknowledge the use of the Casper system (\url{https://ncar.pub/casper}), supported by the NSF National Center for Atmospheric Research (NCAR) at the NSF NCAR-Wyoming Supercomputing Center and sponsored by the National Science Foundation and the State of Wyoming. ChatGPT (OpenAI) was used for language editing to improve clarity and style. The authors reviewed and take full responsibility for the final content of the manuscript.

\clearpage
\bibliographystyle{apalike}
\bibliography{refs}

@article{bruyere2014bias,
  title={Bias corrections of global models for regional climate simulations of high-impact weather},
  author={Bruy{\`e}re, Cindy L and Done, James M and Holland, Greg J and Fredrick, Sherrie},
  journal={Climate Dynamics},
  volume={43},
  number={7},
  pages={1847--1856},
  year={2014},
  publisher={Springer}
}

@article{rahimi2024overview,
  title={An overview of the Western United States dynamically downscaled dataset (WUS-D3)},
  author={Rahimi, Stefan and Huang, Lei and Norris, Jesse and Hall, Alex and Goldenson, Naomi and Krantz, Will and Bass, Benjamin and Thackeray, Chad and Lin, Henry and Chen, Di and others},
  journal={Geoscientific Model Development},
  volume={17},
  number={6},
  pages={2265--2286},
  year={2024},
  publisher={Copernicus GmbH}
}

@article{rahimi2024understanding,
  title={Understanding the cascade: Removing GCM biases improves dynamically downscaled climate projections},
  author={Rahimi, Stefan and Huang, Lei and Norris, Jesse and Hall, Alex and Goldenson, Naomi and Risser, Mark and Feldman, Daniel R and Lebo, Zachary J and Dennis, Eli and Thackeray, Chad},
  journal={Geophysical Research Letters},
  volume={51},
  number={9},
  pages={e2023GL106264},
  year={2024},
  publisher={Wiley Online Library}
}

@article{rodgers2021ubiquity,
  title={Ubiquity of human-induced changes in climate variability},
  author={Rodgers, Keith B and Lee, Sun-Seon and Rosenbloom, Nan and Timmermann, Axel and Danabasoglu, Gokhan and Deser, Clara and Edwards, Jim and Kim, Ji-Eun and Simpson, Isla R and Stein, Karl and others},
  journal={Earth System Dynamics},
  volume={12},
  number={4},
  pages={1393--1411},
  year={2021},
  publisher={Copernicus GmbH}
}

@inproceedings{ronneberger2015u,
  title={U-net: Convolutional networks for biomedical image segmentation},
  author={Ronneberger, Olaf and Fischer, Philipp and Brox, Thomas},
  booktitle={International Conference on Medical image computing and computer-assisted intervention},
  pages={234--241},
  year={2015},
  organization={Springer}
}

@inproceedings{isola2017image,
  title={Image-to-image translation with conditional adversarial networks},
  author={Isola, Phillip and Zhu, Jun-Yan and Zhou, Tinghui and Efros, Alexei A},
  booktitle={Proceedings of the IEEE conference on computer vision and pattern recognition},
  pages={1125--1134},
  year={2017}
}

@article{stengel2020adversarial,
  title={Adversarial super-resolution of climatological wind and solar data},
  author={Stengel, Karen and Glaws, Andrew and Hettinger, Dylan and King, Ryan N},
  journal={Proceedings of the National Academy of Sciences},
  volume={117},
  number={29},
  pages={16805--16815},
  year={2020},
  publisher={National Academy of Sciences}
}

@article{sha2020deepI,
  title={Deep-learning-based gridded downscaling of surface meteorological variables in complex terrain. Part I: Daily maximum and minimum 2-m temperature},
  author={Sha, Yingkai and Gagne II, David John and West, Gregory and Stull, Roland},
  journal={Journal of Applied Meteorology and Climatology},
  volume={59},
  number={12},
  pages={2057--2073},
  year={2020}
}

@article{sha2020deepII,
  title={Deep-learning-based gridded downscaling of surface meteorological variables in complex terrain. Part II: Daily precipitation},
  author={Sha, Yingkai and Gagne II, David John and West, Gregory and Stull, Roland},
  journal={Journal of Applied Meteorology and Climatology},
  volume={59},
  number={12},
  pages={2075--2092},
  year={2020}
}

@article{bano2020configuration,
  title={Configuration and intercomparison of deep learning neural models for statistical downscaling},
  author={Ba{\~n}o-Medina, Jorge and Manzanas, Rodrigo and Guti{\'e}rrez, Jos{\'e} Manuel},
  journal={Geoscientific Model Development},
  volume={13},
  number={4},
  pages={2109--2124},
  year={2020},
  publisher={Copernicus Publications G{\"o}ttingen, Germany}
}

@article{bano2021suitability,
  title={On the suitability of deep convolutional neural networks for continental-wide downscaling of climate change projections},
  author={Ba{\~n}o-Medina, Jorge and Manzanas, Rodrigo and Guti{\'e}rrez, Jos{\'e} Manuel},
  journal={Climate Dynamics},
  volume={57},
  number={11},
  pages={2941--2951},
  year={2021},
  publisher={Springer}
}

@article{maraun2010precipitation,
  title={Precipitation downscaling under climate change: Recent developments to bridge the gap between dynamical models and the end user},
  author={Maraun, Douglas and Wetterhall, Frederick and Ireson, Anderson M and Chandler, Richard E and Kendon, Elizabeth J and Widmann, Martin and Brienen, Stephan and Rust, Henning W and Sauter, Tobias and Theme{\ss}l, Matthias and others},
  journal={Reviews of geophysics},
  volume={48},
  number={3},
  year={2010},
  publisher={Wiley Online Library}
}

@article{danabasoglu2020community,
  title={The community earth system model version 2 (CESM2)},
  author={Danabasoglu, Gokhan and Lamarque, J-F and Bacmeister, J and Bailey, DA and DuVivier, AK and Edwards, Jim and Emmons, LK and Fasullo, John and Garcia, R and Gettelman, Andrew and others},
  journal={Journal of Advances in Modeling Earth Systems},
  volume={12},
  number={2},
  pages={e2019MS001916},
  year={2020},
  publisher={Wiley Online Library}
}

@article{rampal2024enhancing,
  title={Enhancing regional climate downscaling through advances in machine learning},
  author={Rampal, Neelesh and Hobeichi, Sanaa and Gibson, Peter B and Ba{\~n}o-Medina, Jorge and Abramowitz, Gab and Beucler, Tom and Gonz{\'a}lez-Abad, Jose and Chapman, William and Harder, Paula and Guti{\'e}rrez, Jos{\'e} Manuel},
  journal={Artificial Intelligence for the Earth Systems},
  volume={3},
  number={2},
  pages={230066},
  year={2024},
  publisher={American Meteorological Society}
}

@article{hernanz2024limitations,
  title={On the limitations of deep learning for statistical downscaling of climate change projections: The transferability and the extrapolation issues},
  author={Hernanz, Alfonso and Correa, Carlos and S{\'a}nchez-Perrino, Juan-Carlos and Prieto-Rico, Ignacio and Rodr{\'\i}guez-Guisado, Esteban and Dom{\'\i}nguez, Marta and Rodr{\'\i}guez-Camino, Ernesto},
  journal={Atmospheric Science Letters},
  volume={25},
  number={2},
  pages={e1195},
  year={2024},
  publisher={Wiley Online Library}
}

@article{doury2023regional,
  title={Regional climate model emulator based on deep learning: Concept and first evaluation of a novel hybrid downscaling approach},
  author={Doury, Antoine and Somot, Samuel and Gadat, Sebastien and Ribes, Aur{\'e}lien and Corre, Lola},
  journal={Climate Dynamics},
  volume={60},
  number={5},
  pages={1751--1779},
  year={2023},
  publisher={Springer}
}

@article{goodfellow2014generative,
  title={Generative adversarial nets},
  author={Goodfellow, Ian J and Pouget-Abadie, Jean and Mirza, Mehdi and Xu, Bing and Warde-Farley, David and Ozair, Sherjil and Courville, Aaron and Bengio, Yoshua},
  journal={Advances in neural information processing systems},
  volume={27},
  year={2014}
}

@article{kouw2019review,
  title={A review of domain adaptation without target labels},
  author={Kouw, Wouter M and Loog, Marco},
  journal={IEEE transactions on pattern analysis and machine intelligence},
  volume={43},
  number={3},
  pages={766--785},
  year={2019},
  publisher={IEEE}
}

@article{hess2025fast,
  title={Fast, scale-adaptive and uncertainty-aware downscaling of Earth system model fields with generative machine learning},
  author={Hess, Philipp and Aich, Michael and Pan, Baoxiang and Boers, Niklas},
  journal={Nature Machine Intelligence},
  volume={7},
  number={3},
  pages={363--373},
  year={2025},
  publisher={Nature Publishing Group UK London}
}

@article{maraun2015value,
  title={VALUE: A framework to validate downscaling approaches for climate change studies},
  author={Maraun, Douglas and Widmann, Martin and Guti{\'e}rrez, Jos{\'e} M and Kotlarski, Sven and Chandler, Richard E and Hertig, Elke and Wibig, Joanna and Huth, Radan and Wilcke, Renate AI},
  journal={Earth's Future},
  volume={3},
  number={1},
  pages={1--14},
  year={2015},
  publisher={Wiley Online Library}
}

@article{hawkins2012time,
  title={Time of emergence of climate signals},
  author={Hawkins, Ed and Sutton, Rowan},
  journal={Geophysical Research Letters},
  volume={39},
  number={1},
  pages={L01702},
  year={2012},
  doi={10.1029/2011GL050087},
  publisher={Wiley Online Library}
}

@article{zhang2011indices,
  title={Indices for monitoring changes in extremes based on daily temperature and precipitation data},
  author={Zhang, Xuebin and Alexander, Lisa and Hegerl, Gabriele C and Jones, Philip and Tank, Albert Klein and Peterson, Thomas C and Trewin, Blair and Zwiers, Francis W},
  journal={Wiley Interdisciplinary Reviews: Climate Change},
  volume={2},
  number={6},
  pages={851--870},
  year={2011},
  publisher={Wiley Online Library}
}

\clearpage
\section*{Supplementary Information}
\setcounter{table}{0}
\setcounter{figure}{0}
\renewcommand{\thetable}{S\arabic{table}}
\renewcommand{\thefigure}{S\arabic{figure}}

\subsection*{Model Details}
\label{sec:model_details}
In this subsection, we provide a formal specification of the generator-discriminator architecture used in the main experiments. Let \(x\in\mathbb{R}^{H\times W\times C_{\mathrm{in}}}\) denote the resized multivariate climate input and let \(\hat{y}=G_{\theta}(x)\in\mathbb{R}^{H\times W\times C_{\mathrm{out}}}\) denote the predicted high-resolution field, where \(C_{\mathrm{in}}\) is the number of input channels and \(C_{\mathrm{out}}\) is the number of target channels. Here, each atmospheric variable, pressure level, and vector component is represented as a separate channel. We use the same architecture and training configuration for all temporal sampling strategies so that differences in downstream performance can be attributed to data selection rather than changes in model design.

\subsubsection*{Generator architecture}
The generator \(G_{\theta}\) follows a U-Net-style encoder-decoder architecture. It accepts a multivariate climate input with \(C_{\mathrm{in}}\) channels and predicts \(C_{\mathrm{out}}\) output channels. The encoder is organized as a feature pyramid with channel widths
\[
  c_0=b,\qquad c_1=2b,\qquad c_2=4b,\qquad c_3=8b,\qquad c_4=968,
\]
where the base width is \(b=88\). This value was chosen to yield well-proportioned channel sizes across the network relative to the input-channel dimensionality, the target bottleneck width, the use of Group Normalization with 8 groups, and the decoder's transposed-convolution (deconvolution) upsampling path. Thus, the successive encoder stages use widths \(88\), \(176\), \(352\), \(704\), and \(968\). Each stage uses replicate-padded convolutions, Group Normalization with 8 groups, and SiLU nonlinearities. The first encoder stage preserves the input spatial resolution, while the next four stages downsample via stride-2 convolutions. To regularize the deepest representations without perturbing the shallower multiscale features, Dropout2d with probability \(p=0.1\) is applied only in the last two encoder stages.

The bottleneck consists of two replicate-padded \(3\times 3\) convolutions at 968 channels. Each convolution is preceded by GroupNorm and a SiLU activation, and each is followed by Dropout2d with \(p=0.1\). This bottleneck preserves the deepest channel width while increasing nonlinear mixing at the coarsest spatial scale.

The decoder mirrors the multiscale structure of the encoder. Each decoder stage upsamples its incoming features using a transposed-convolution block, concatenates the upsampled representation with the corresponding encoder activations through standard U-Net skip connections, and refines the combined tensor using GroupNorm, SiLU activations, and additional convolutions. Dropout is retained in the two deepest decoder stages and omitted in the shallower stages. The output head uses a residual tail block composed of GroupNorm, SiLU, and two \(1\times 1\) convolutions, followed by a final \(1\times 1\) projection that maps the hidden representation to the required number of target channels \(C_{\mathrm{out}}\).

\subsubsection*{Discriminator and training}
The adversarial discriminator \(D_{\phi}\) is a PatchGAN-style network tailored to generated climate fields. It takes either a predicted output or a reference target field with \(C_{\mathrm{out}}\) channels as input and processes it through four convolutional blocks with kernel size \(4\times 4\), stride 2, and padding 1. The discriminator channel widths are 64, 128, 256, and 512. Each intermediate block uses a LeakyReLU activation with negative slope 0.2. No normalization layers or dropout are used in the discriminator. A final \(4\times 4\) convolution with stride 1 produces a single-channel patch score map, which supplies the adversarial signal at the level of local spatial patches.

All models are trained under a common optimization setup to isolate the effect of training data selection. The primary reconstruction objective is mean absolute error (MAE/L1),
\[
  \mathcal{L}_{\mathrm{rec}} = \lVert \hat{y} - y \rVert_1,
\]
which is applied consistently across all runs. The adversarial component implemented with Softplus-based GAN losse. Using a fixed generator, discriminator, and loss configuration across experiments ensures that comparisons remain data-centric and are not confounded by changes in architecture or objective design.

Training follows the same preprocessing and normalization pipeline described in the main text, including split-aware scaling and evaluation on held-out ensemble runs. Checkpoint selection is based on validation behavior under the corresponding training subset, and models are trained long enough to reach stable performance under each subsampling setting. This shared protocol ensures that variation in results reflects differences in temporal coverage and subsampling level, rather than inconsistent training procedures.


\begin{table}[!ht]
  \centering
  \caption{Input predictor variables and units used in this study. For 3-D atmospheric components (\texttt{t\_3d}, \texttt{u\_3d}, \texttt{v\_3d}), values are taken at 850, 700, 500, and 300 hPa pressure levels.}
  \label{tab:si_input_predictors}
  \begin{tabular}{l p{0.58\linewidth}}
    \hline
    \textbf{Variable} & \textbf{Description and Units} \\
    \hline
    \texttt{cape} & Convective available potential energy [J kg$^{-1}$] \\
    \texttt{ivt} & Integrated vapor transport (zonal and meridional components; earth relative) [kg s$^{-1}$ m$^{-1}$] \\
    \texttt{prec} & Precipitation rate [mm d$^{-1}$] \\
    \texttt{q2} & 2-m specific humidity [kg kg$^{-1}$] \\
    \texttt{snow} & Snow water equivalent [mm] \\
    \texttt{t2} & 2-m average temperature [K] \\
    \texttt{t\_3d} & 3-D temperature [K] \\
    \texttt{u\_3d} & 3-D \texttt{u} component of wind (earth relative) [m s$^{-1}$] \\
    \texttt{v\_3d} & 3-D \texttt{v} component of wind (earth relative) [m s$^{-1}$] \\
    \texttt{uv10} & 10-m \texttt{u}, \texttt{v} (earth relative) [m s$^{-1}$] \\
    \hline
  \end{tabular}
\end{table}

\begin{figure}[h!]
  \centering
  \includegraphics[width=0.99\linewidth]{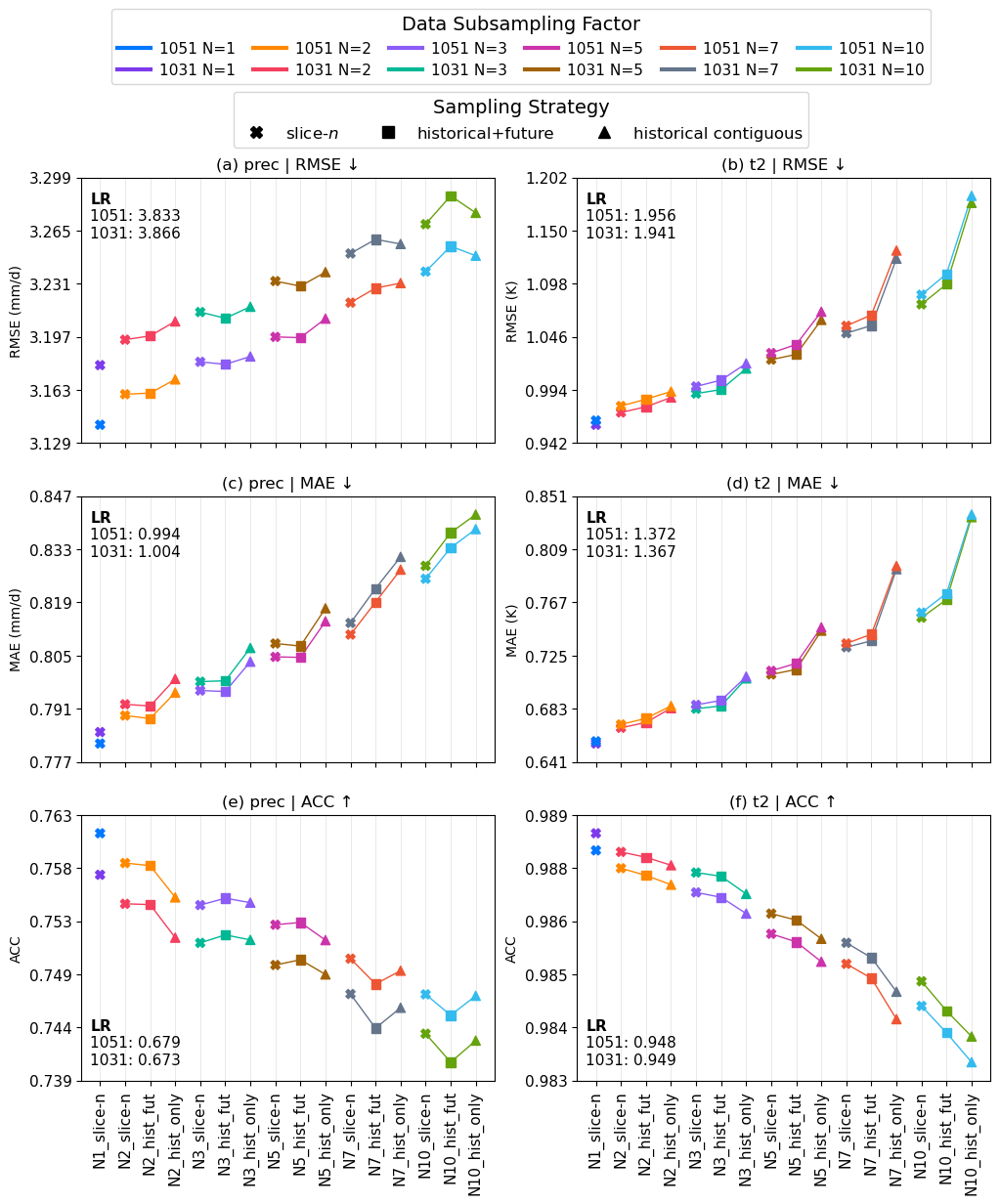}
  \caption{\small{Model performance across data subsampling factors and sampling strategies for two CESM2-LE realizations, 1031 and 1051. Panels show RMSE, MAE, and ACC for \texttt{prec} and \texttt{t2}. Colors indicate the data subsampling factor, with separate colors for each CESM2-LE realization, and markers indicate the sampling strategy. Across both realizations, performance changes consistently with increasing subsampling factor, while differences between the 1031 and 1051 runs remain small. This indicates that the relative effects of subsampling factor and sampling strategy are robust across CESM2-LE realizations. Lower RMSE and MAE indicate better performance, while higher ACC indicates better performance.}}
  \label{fig:1031_vs_1051_pointwise_stats}
\end{figure}

\end{document}